\documentclass[sigconf,screen,nonacm]{acmart}

\usepackage{tabularx}
\usepackage{booktabs}
\usepackage{xcolor}
\usepackage{bm}
\usepackage{caption}

\usepackage{subfigure}

\AtBeginDocument{%
  \providecommand\BibTeX{{%
    \normalfont B\kern-0.5em{\scshape i\kern-0.25em b}\kern-0.8em\TeX}}}

\makeatletter
\def\runningfoot{\def\@runningfoot{}}
\def\firstfoot{\def\@firstfoot{}}
\makeatother

\newcommand{\updates}[1]{\color{black}  #1}

\renewcommand\footnotetextcopyrightpermission[1]{}
\setcopyright{acmcopyright}
\copyrightyear{2022}
\acmYear{2022}
\acmDOI{}

\acmConference[ACMMM 2022]{Make sure to enter the correct
  conference title from your rights confirmation emai}{Oct 10--14,
  2022}{Lisbon, Portugal}
\acmPrice{}
\acmISBN{}

\begin{document}

\title{Self-Supervised Video Object Segmentation via Cutout Prediction and Tagging}

\author{Jyoti Kini}
\email{jyoti.kini@knights.ucf.edu}
\affiliation{%
  \institution{University of Central Florida}
  \country{USA}
}

\author{Fahad Shahbaz Khan}
\email{fahad.khan@liu.se}
\affiliation{%
  \institution{MBZ University of Artificial Intelligence$^{1}$}
  \institution{Linköping University$^{2}$}
  \country{UAE$^{1}$, Sweden$^{2}$}
}

\author{Salman Khan}
\email{salman.khan@mbzuai.ac.ae}
\affiliation{%
  \institution{MBZ University of Artificial Intelligence}
  \country{UAE}
}

\author{Mubarak Shah}
\email{shah@crcv.ucf.edu}
\affiliation{%
  \institution{University of Central Florida}
  \country{USA}
}

\begin{abstract}
We propose a novel self-supervised Video Object Segmentation (VOS) approach that strives to achieve better object-background discriminability for accurate object segmentation. Distinct from previous self-supervised VOS methods, our approach is based on a discriminative learning loss formulation that takes into account both object and background information to ensure object-background discriminability, rather than using only object appearance. The discriminative learning loss comprises cutout-based reconstruction (cutout region represents part of a frame, whose pixels are replaced with some constant values) and tag prediction loss terms. The cutout-based reconstruction term utilizes a simple cutout scheme to learn the pixel-wise correspondence between the current and previous frames in order to reconstruct the original current frame with added cutout region in it. The introduced cutout patch guides the model to focus as much on the significant features of the object of interest as the less significant ones, thereby implicitly equipping the model to address occlusion-based scenarios. Next, the tag prediction term encourages object-background separability by grouping tags of all pixels in the cutout region that are similar, while separating them from the tags of the rest of the reconstructed frame pixels. Additionally, we introduce a zoom-in scheme that addresses the problem of small object segmentation by capturing fine structural information at multiple scales. Our proposed approach, termed  CT-VOS, achieves state-of-the-art results on two challenging benchmarks: DAVIS-2017 and Youtube-VOS. A detailed ablation showcases the importance of the proposed loss formulation to effectively capture object-background discriminability and the impact of our zoom-in scheme to accurately segment small-sized objects.  
\end{abstract}

\keywords{self-supervision, video object segmentation, cutout, tagging, pretext tasks, attention}

\maketitle
\pagestyle{plain}

\section{Introduction}
\label{sec:intro}
Video object segmentation (VOS) aims to segment an object of interest in a video, given its segmentation mask in the first frame. VOS plays an important role in numerous real-world applications, such as interactive video editing, autonomous driving, augmented-reality, and smart surveillance systems. The problem is challenging since the target object is only provided with a reference segmentation in the first frame, and determining the object segmentation mask in all other frames involves: {\bf a.} accounting for drastic changes in the appearance of the object of interest, over time, due to occlusion, {\bf b.} distinguishing the object from highly similar surrounding distractor elements, and {\bf c.} estimating finer object details, especially in case of a small object or object diminishing in size across the video sequence.

\begin{figure}[t]
\centering
\includegraphics[height=4.8cm, width=8.4cm]{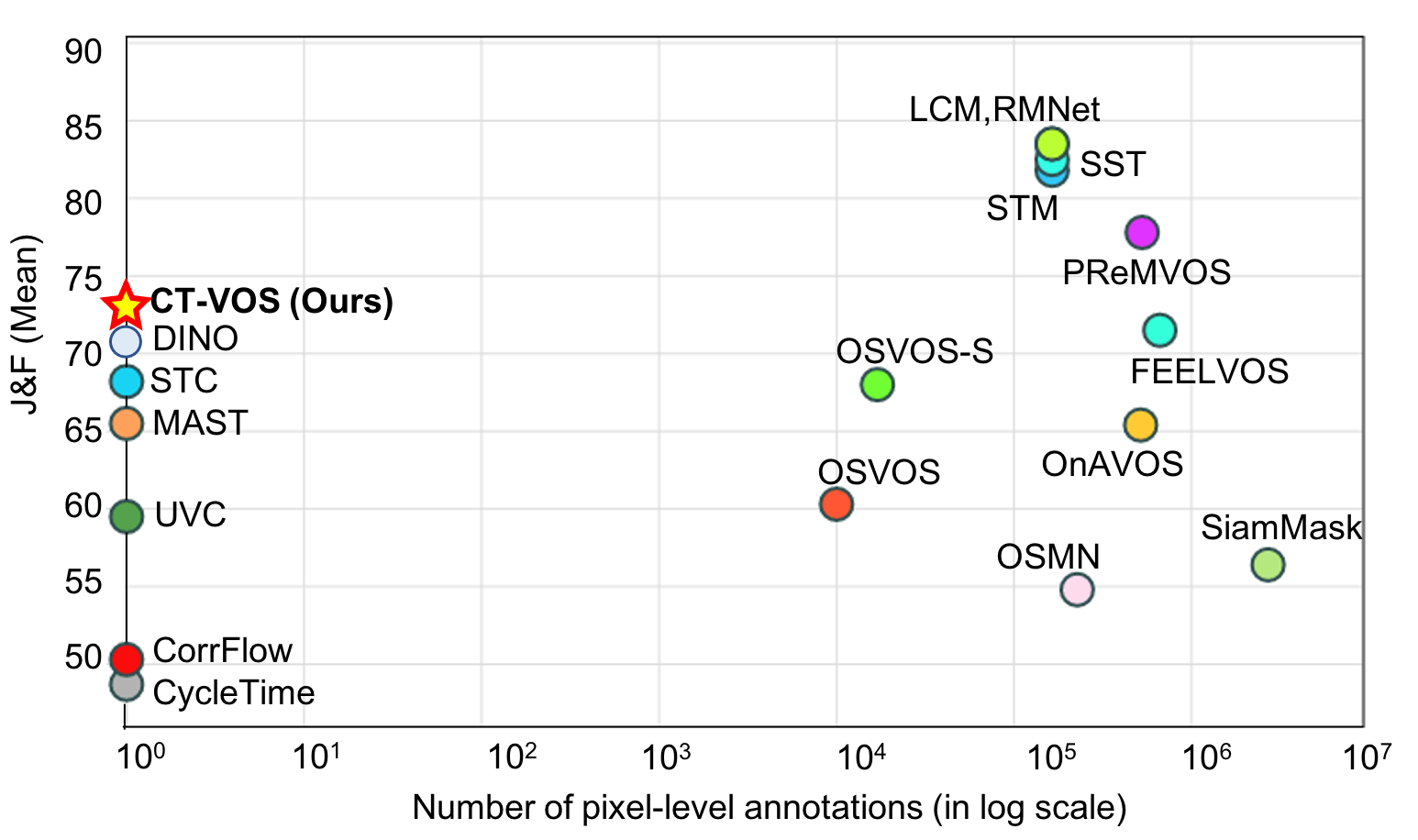}
\caption{Comparison in terms of overall accuracy $J\&F (Mean)$ and pixel-level annotations required for different VOS methods on DAVIS-2017. The proposed CT-VOS outperforms existing self-supervised VOS methods. Different VOS methods shown here in comparison are:  DINO \cite{Caron_2021_ICCV}, STC \cite{NEURIPS2020_e2ef524f}, MAST \cite{lai2020mast}, UVC \cite{li2019joint}, CorrFlow \cite{lai2019self}, CycleTime \cite{wang2019learning}, LCM \cite{hu2021learning}, RMNet \cite{Xie_2021_CVPR}, SST \cite{duke2021sstvos}, STM \cite{oh2019video}, PReMVOS \cite{luiten2018premvos}, FEELVOS \cite{voigtlaender2019feelvos}, OSVOS\_S \cite{maninis2018video}, OnAVOS \cite{DBLP:conf/bmvc/VoigtlaenderL17}, OSVOS \cite{Cae+17}, SiamMask \cite{wang2019fast}, OSMN \cite{8578778}.
}
\label{fig:Teaser}
\end{figure}

Most existing VOS approaches \cite{5bdc315017c44a1f58a05df8, oh2019video, Duarte_2019_ICCV, conf/cvpr/ChenPMG18, Voigtlaender19CVPR, 8237742, 8099855} typically rely on extensive supervision during training in the form of ground-truth human annotations (see Figure \ref{fig:Teaser}). Specifically, a common strategy is to fine-tune a network backbone, conventionally pre-trained on ImageNet \cite{deng2009imagenet}, using additional data (\emph{e.g.} MS COCO \cite{lin2014microsoft}, Pascal VOC \cite{everingham2010pascal}, DAVIS \cite{pont20172017}, YouTube-VOS \cite{xu2018youtube}) with pixel-level supervision in the form of segmentation masks. Different from these supervised approaches, recent self-supervised VOS methods \cite{lai2020mast, zhu2020self, yang2021self, wang2019learning, lai2019self, vondrick2018tracking} perform segmentation without utilizing additional supervised training in the form of pixel-level ground-truth annotations. 
Several of these self-supervised approaches learn pixel-wise correspondence between frames by reducing the reconstruction error \cite{lai2020mast, vondrick2018tracking}, whereas others focus on matching pixels across frames using inherent cyclic consistency in videos \cite{wang2019learning, lai2019self}. While promising
results have been achieved by these self-supervised approaches, the performance still remains far from their supervised counterparts.

Typically, self-supervised VOS \cite{lai2020mast} approach employs robust pixel matching strategies to learn prominent representations of the target region while paying minimal focus on learning the feature embedding of the background region. Our methodology, however, is based on introducing feature diversity by taking into account both object and background information for object-background discriminability. The proposed cutout and tagging learning scheme encourages the object prediction model to focus on both prominent and non-prominent features of the target object as well as the background appearance, providing superior discriminative abilities. In addition, we alleviate a common problem encountered in several existing VOS methodologies that of loss of details, especially in small objects.

We introduce a self-supervised VOS approach, CT-VOS, that predicts the segmentation mask for the current frame based on the information in previous frames. Instead of learning trivial solutions from standard self-supervised color-based reconstruction loss between the current frame and previous frames, we utilize a two-fold discriminative learning loss formulation. \emph{First,} instead of simply reconstructing the current frame, we also predict cutout in the current frame based on cutout in previous frames. Here, cutout represents a part of an image, whose pixels are replaced with some constant values \emph{e.g.}, zero. Intuitively, cutouts effectively simulate occlusion-based discontinuities and thereby aid the model to deal with occlusions during inference. \emph{Second,} the model predicts tags in the current frame, such that the tags of all pixels in the cutout region are similar, and also considerably different from tags of non-cutout pixels. Here, we generate a heatmap of per-pixel identity tags for the current frame, where each tag could be any arbitrary value between 0 and 1. Our tagging loss enables separating the cutout pixels from the reconstructed image pixels, thereby improving the VOS performance by mimicking foreground-background separation. These complementary formulations are designed to allow the model to learn a rich representation, rather than learning a trivial solution of assigning the exact same color to pixels. Furthermore, we utilize zoom-in views of a random frame that facilitates the model to deal with objects at different scales.
To summarize, our contribution is multifold: 
\begin{itemize}
  \item[$\bullet$] 
  We present an object-background discriminative loss formulation comprising a cutout-based reconstruction element and tagging loss component. While cutout-based reconstruction forces the model to focus on both salient as well as less salient aspects of the object of interest by simulating controlled spatio-temporal occlusion-based discontinuities, tagging loss effectively separates the object of interest from the background pixels by mimicking foreground-background separation.
  \item[$\bullet$] We introduce zoom-in views in our VOS architecture that enables processing small spatial regions and thereby captures fine-grained object details. The zoom-in scheme not only assists with tracking smaller objects but also improves the segmentation mask details generated for larger objects.
  \item[$\bullet$] Comprehensive qualitative and quantitative experiments are performed on two challenging benchmarks: DAVIS-2017 \cite{pont20172017} and Youtube-VOS \cite{xu2018youtube}. Our results reveal the benefits of object-background discriminative loss formulation and zoom-in scheme, outperforming all published self-supervised VOS methods on both benchmarks.
\end{itemize}

\begin{figure*}[ht]
\begin{center}
\includegraphics[height=7.5cm, width=14cm]{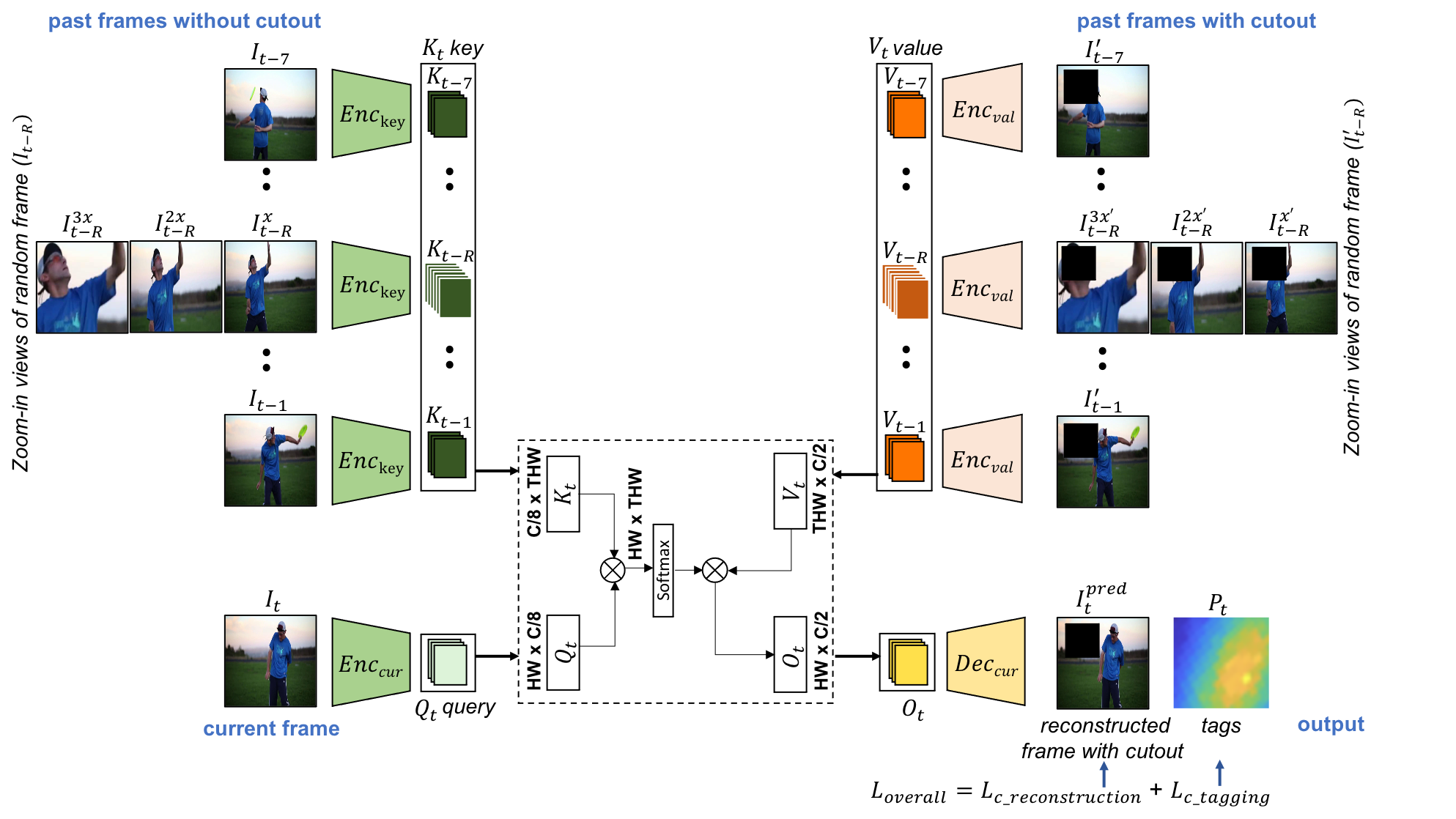}
\end{center}\vspace{-1em}
   \caption{Overview of the proposed CT-VOS training. Firstly, the current frame $I_t$ is used to generate query $Q_{t}$, and the past frames $I_{t\it{-1}}, $\ldots$, I_{t\it{-7}}$ are employed to compute a comprehensive key set $\bm{K}_{t}$ (shown on the left). Next, the past frames with cutout $I^{'}_{t\it{-1}}, $\ldots$, I^{'}_{t\it{-7}}$ are processed to yield the value set $\bm{V}_{t}$ (shown on the right). Then, the generated query, key and value are employed to predict the reconstructed frame with cutout  $I_{t}^{pred}$ and  tag channel $P_t$ (shown on the bottom right). In order to learn finer object details, we introduce zoom-based strategy by choosing a random frame $I_{t-R}$ (shown on the left). Here, we introduce 3 different zoom levels in a chosen random frame. Therefore, with these three frames, the total number of frames, $T$,  corresponds to 10 frames. $\otimes$ implies dot product.}
\label{fig:Architecture}
\end{figure*}

\section{Related Work}
Hand-crafted features based on appearance, motion, spatial proximity, and neighbourhood grouping constraints have been used in earlier works \cite{10.1007/978-3-642-15555-0_21, BMVC.28.21, 10.1007/978-3-319-10593-2_43, NSB15} on VOS. The need for extensive manual annotations in a supervised setting has encouraged researchers to venture beyond the supervised paradigm into the realm of self-supervised VOS. Based on the nature of supervision, existing VOS approaches can be roughly categorised into following three groups:

\noindent \textbf{Supervised VOS Methods:}
A considerable volume of work on video object segmentation (VOS) task \cite{5bdc315017c44a1f58a05df8, oh2019video, Duarte_2019_ICCV, conf/cvpr/ChenPMG18, Voigtlaender19CVPR} focuses on segmenting each object instance separately. Several recent works \cite{5bdc315017c44a1f58a05df8, Duarte_2019_ICCV, 8237742}, employ sequential modelling to achieve temporal consistency. Lately, a number of solutions have been motivated by embedding based approaches \cite{8099855, Voigtlaender19CVPR}, which deal with learning pixel-mappings between reference and target frames. One such prior work by Oh et al. \cite{oh2019video}, deals with binary-classification of each pixel to generate a single-object segmentation mask for a given frame by applying non-local matching with previously predicted frames. Another idea of the region based segmentation approach is used to extract multiple object proposals in a single feed forward-path, followed by associating objects across frames \cite{li2018video, luiten2018premvos}. However, the dependence of these approaches on region proposal networks \cite{NIPS2015_5638, 8237584} and ground-truth annotations introduces additional complexities. 

\noindent \textbf{Few-Shot Learning Based Segmentation:}
Few-shot segmentation deals with learning to segment objects based on a limited set of annotated examples. Initial literature \cite{shaban2017one, dong2018few, siam2019amp, zhang2019pyramid} in this area focused on using labeled segmentation masks for few examples in the support set. Some of the latest methods \cite{zhang2019canet, rakelly2018conditional, wang2019panet} have worked with weaker annotations based on scribble or bounding boxes. Raza et al. \cite{raza2019weakly} recently proposed using image-level labels to solve the supervision requirement in a few-shot segmentation task. Although this is a promising direction, these formulations are labor-intensive, especially in robotic applications that require learning online.

\noindent \textbf{Self-supervised Segmentation:}
Self-supervised methods enable models to learn without manually generated supervisory signals, thereby eliminating the overhead of large annotated datasets. Generally, computer vision pipelines that employ self-supervised learning employ two tasks: a pretext task and a downstream task. Although the downstream task is the ultimate goal, the pretext task is the primary task that the self-supervised method seeks to solve. In the literature, some pretext tasks have been explored to learn spatial context from images and videos \cite{doersch2015unsupervised, lai2020mast}, while others have relied on the temporal ordering and consistency \cite{kim2019self}.  Generation of pseudo classification labels \cite{komodakis2018unsupervised}, future representations prediction  \cite{han2019video, vondrick2016anticipating}, jigsaw puzzles prediction \cite{noroozi2016unsupervised}, arrow of time \cite{wei2018learning} and shuffled frames \cite{misra2016shuffle} are some other popular pretext tasks. Recent work by Han et al. \cite{han2019video} uses contrastive learning on videos to learn strong video representations. In  CT-VOS, instead of learning trivial solutions from standard self-supervised color-based reconstruction loss between the current frame and previous frames, we employ two-fold object-background discriminative loss formulation and a zoom-in scheme for self-supervision. 

\section{Approach}
Given the first frame segmentation mask of an object, the proposed network segments the object throughout the video clip based on the context provided by the previous frames. Figure \ref{fig:Architecture} illustrates the overall network architecture. We exploit the automatically available supervisory signals from data by framing VOS as a self-supervised learning problem. To this end, we introduce object-background discriminative loss formulation comprising cutout-based reconstruction and tag (cutout vs. non-cutout) prediction. While one of the components helps mimic occlusion-based spatio-temporal discontinues, the other encourages the model to generate well-separated foreground-background segmentation masks. We also utilize zoom-in views of a random frame that enriches the overall representation by capturing  the segmentation details at different scales.

Initially, the current frame is processed through an encoder setup to generate {\em current frame embedding - query} (Figure \ref{fig:Architecture}). Next, previous frames are encoded to {\em past frame embeddings - keys}. The query interacts with keys to generate similarities between the current and previous frame content. Then, the {\em values} are extracted using previous frames with introduced cutout patches. Finally, the relative matching scores and values are fed to the {\em decoder} to generate the reconstructed image with cutout prediction and a tag map. During inference stage, the values are generated using previously predicted object segmentation masks, and the decoder yields object segmentation mask along with the tags of the current frame. 

\subsection{Architecture}
\label{Architecture}
\begin{figure}[!t]
\centering
\includegraphics[height=3.8cm, width=9cm]{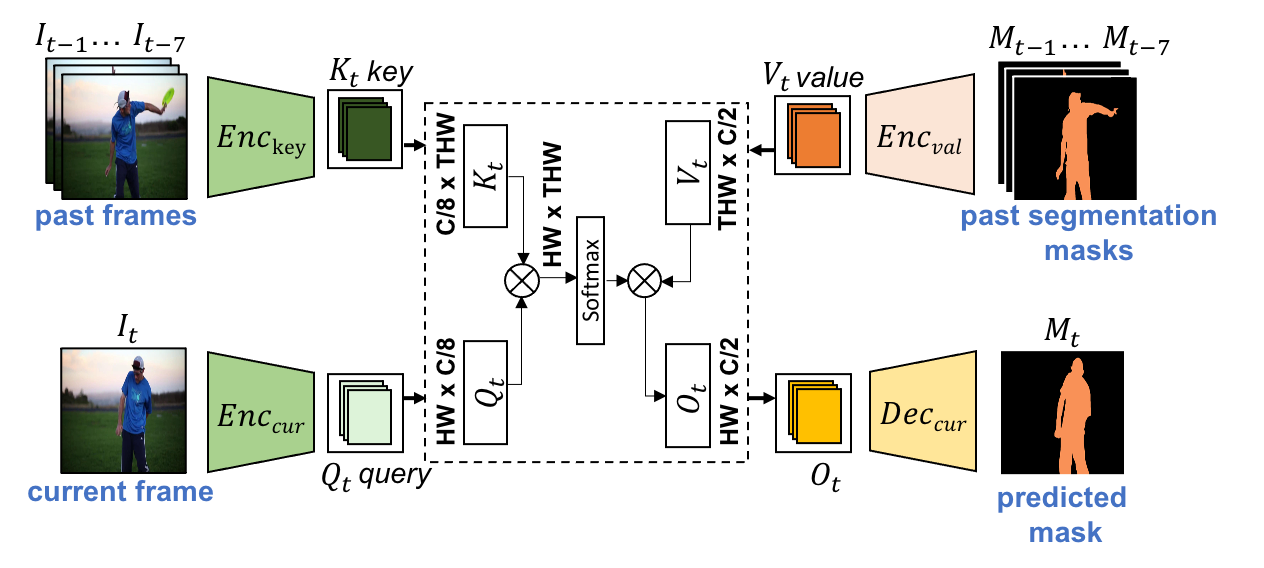}
\vspace{-1em}
\caption{\emph{Inference setup:} During testing phase, the value $\bm{V}_{t}$ is computed using previous segmentation masks (shown on the top right). The decoder generates object segmentation mask $M_t$ as output. Here, $\bm{V}_{t}$ is generated by repeating each segmentation mask across 3 channels, and $M_t$ is resultant single channel output by averaging the 3 channel output from the decoder normalised to range [-1, 1] that is further thresholded at 0. }
\label{fig:Inference}
\end{figure}

Here, we elaborate the preliminaries of our network presented in Figure \ref{fig:Architecture} and Figure \ref{fig:Inference}. Using a ResNet-18 \cite{he2016deep} backbone network $Enc_{cur}$, we encode the current frame $I_t$ into current frame embedding or query $Q_{t}$. Next, the past frames $I_{t-1}$...$I_{t-7}$ are encoded by $Enc_{key}$ to generate keys $K_{t-1}$...$K_{t-7}$. These keys are concatenated to form a comprehensive key $\bm{K}_{t}$. Using the query and the key, attention is computed, which is used to determine the strength of the correspondence between pixels from the current and the past frames. During training, past frames introduced with cutouts $I_{t-1}^{'}$...$I_{t-7}^{'}$ are passed through $Enc_{val}$. 
The resultant generated values $V_{t-1}$...$V_{t-7}$ (collectively termed as $\bm{V}_{t}$) are then combined with the query and the key. We utilize this resultant representation as input to the decoder $Dec_{cur}$, which yields the reconstructed image with cutout  $I_{t}^{pred}$ and tags $P_t$. 

The model comprises ResNet-18 as the backbone network for current frame encoder $Enc_{cur}$,  past frames encoder $Enc_{key}$ as well as value encoder $Enc_{val}$. The base features are computed using stage-4 (res4) of the ResNet-18 model. For each input frame $I_t$, we have a  triplet ({$Q_t$, $\bm{K}_t$, $\bm{V}_t$}) referring to Query, Key, and Value. During training, reconstruction of the cutout-based current frame ($I_{t}^{pred}$) is accomplished by  combining pixels from the current frame ($I_{t}$), with  pixels of past frames through attention.   

To implement cutout, we apply a fixed-size zero mask to a random location in the frame. The selected random location for the cutout is retained for each image within a 10-frame clip passed through $Enc_{val}$, as shown in Figure \ref{fig:Architecture}. In addition, we apply a cutout size constrain such that at least 50\% of the image remains unmodified.
Concurrently, we generate multi-scale feature embeddings for one of the  random past frames. This helps in  dealing with multiple object sizes and finer deals, thereby enhancing the robustness of the method.

During inference, however, as illustrated in Figure \ref{fig:Inference}, the approach focuses on generating object segmentation mask ${M}_{t}$ for time frame $t$ (current frame). While the network is trained on $\bm{V}_{t}$ comprising 3-channel cutout-based RGB frames, we generate 3-channel $\bm{V}_{t}$ from 1-channel segmentation masks $M_{t-1}$...$M_{t-7}$ during inference by stacking each object mask 3 times. Also, during training, the resultant output $\bm{O}_{t}$ is a cutout-based reconstructed RGB frame, therefore inference involves computing resultant 1-channel segmentation output $M_t$ by averaging 3-channel output from the decoder $Dec_{cur}$ normalized to the range [-1, 1] and further thresholded at 0. Additionally, we do not use the tags generated during inference.

\subsection{Proposed Loss Formulation}
\label{Training Losses}
To train our network, we introduce two new losses, namely self-supervised cutout-based reconstruction loss and tagging loss, detailed below:
\begin{align}
    L_{overall} = L_{c\_reconstruction} + \lambda L_{c\_tagging}.
\end{align}
The reconstruction loss with cutout ensures that the model accounts for the less significant features within the frames as much as it focuses on the salient features. This better equips the VOS model with occlusion-based spatio-temporal discontinuities that it would encounter during inference. In addition, the self-supervised tagging loss enforces the separation of the object from the background to boost the segmentation performance. We use a combination of cutout-based reconstruction loss and $\lambda$ weighted self-supervised tagging loss to train our network for segmentation (equally weighted losses give us the best results). 

\noindent\textbf{Reconstruction Loss with Cutout:}
The objective $L_{c\_reconstruction}$  to distinguish between the reconstructed frame $I_{t}^{pred}$ and input frame $I_{t}$ (we are dropping subscript $t$ below for convenience), is defined in terms of Huber Loss as follows:
\begin{align}
    L_{c\_reconstruction} = \frac{1}{{N}}\sum_{j=1}^{N}{z_{j}},
\end{align}
\begin{align}
    \text{ where }  
    z_{j} = \begin{cases}
                0.5(I_{j}^{pred} - I_{j})^{2} & if 
                \ $\textbar$ I_{j}^{pred} - I_{j} 
                $\textbar$ \
                < 1\\
               |I_{j}^{pred} - I_{j}| - 0.5 & otherwise
            \end{cases}
\end{align}
\noindent Here, $N$ corresponds to total  number of pixels in the image, and  we assume pixel values in $I$ and $I^{pred}$ are normalized between -1 and 1. 

\noindent\textbf{Self-Supervised Cutout-Based Tagging Loss:}
We introduce the self-supervised cutout-based tagging loss that comprises a cutout pixel pull loss, $L_{cp}$, and cutout pixel-to-reconstructed image pixel push loss, $L_{rp}$, for the separation of pixels in the VOS task. Suppose, $h_{w}$ is the predicted tag (or the regressed value, i.e. continuous values between 0 and 1) for a given $w^{th}$ random pixel in the cutout region or the reconstructed image region, then let us define $h_{\updates W}^{'}$, that is the mean of tags for the group of $W$ {\updates random} pixels as:
\begin{align}
    h_{\updates W}^{'} = \frac{1}{W}\sum_{w=1}^{W}{h_{w}}.
\end{align}
{\updates In order to define $L_{cp}$, we only work with randomly selected subset of cutout pixels, say $m$ where $m = 1,...M$. Here, $h_m$ corresponds to predicted tag for $m^{th}$ cutout pixel, while $h_M^{'}$ refers to mean of tags of M cutout pixels.} 
Formally, we define $L_{cp}$ below which pulls the tags of all cutout pixels closer as:
\begin{align}
    L_{cp} = \frac{1}{{\updates M}}\sum_{{\updates m}=1}^{{\updates M}}{(h_{\updates m} - h_{\updates M}^{'})^{2}}.
\end{align}
In addition to pulling the tags of cutout pixels together, we also try to push apart the cutout pixels from the reconstructed image pixels. We introduce a margin $G$ to provide a permissible gap on the difference in the embedding space of cutout pixels ${\updates C}$ and reconstructed image pixels ${\updates B}$. {\updates Given the  mean of cutout pixel tags $h_{C}^{'}$} and the mean of reconstructed image pixel tags $h_{B}^{'}$, the complementary margin-based penalty loss $L_{rp}$ is given by:  
\begin{align}
    L_{rp} = max(0,G-{||h_{\updates B}^{'} - h_{\updates C}^{'}||}).
\end{align}
The self-supervised cutout-based tagging loss $L_{c\_tagging}$ that helps with separation of cutout pixels and reconstructed image pixels is defined as: 
\begin{align}
    L_{c\_tagging} = L_{cp} + L_{rp}.
\end{align}

\setlength{\tabcolsep}{0.8pt}
\begin{table}[t]
    \begin{center}
        \renewcommand{\arraystretch}{1}
        \begin{tabularx}{8cm}{@{}l*{7}{c}c@{}}
        \toprule
        {Method} & 
        {Backbone} &  
        {Dataset (Size)} & 
        {\emph{$J$\&F}} &
        {\emph{$J$\textsubscript{mean}}} & 
        {\emph{F\textsubscript{mean}}} \\
        \midrule
        {\bf Unsupervised} \\
        \midrule
        Vid. Color \cite{vondrick2018tracking} & ResNet-18 & Kinetics (800 hrs) & 34.0 & 34.6 & 32.7 \\
        CycleTime \cite{wang2019learning} & ResNet-50 & VLOG (344 hrs) & 48.7 & 46.4 & 50.0 \\
        CorrFlow \cite{lai2019self} & ResNet-18 & OxUvA (14 hrs) & 50.3 & 48.4 & 52.2 \\
        UVC \cite{li2019joint} & ResNet-18 & Kinetics (800 hrs) & 59.5 & 57.7 & 61.3 \\
        MAST \cite{lai2020mast} & ResNet-18 & YT-VOS (5.58 hrs) & 65.5 & 63.3 & 67.6 \\
        {\bf CT-VOS (Ours)} & ResNet-18 & YT-VOS (5.58 hrs) & {\bf 68.2} & {\bf 67.1} & {\bf 69.2} \\
        \updates{STC \cite{NEURIPS2020_e2ef524f}} & \updates{ResNet-18} & 
        \updates{Kinetics (800 hrs)} & 
        \updates{68.3} & 
        \updates{65.5} & 
        \updates{71.0} \\
        \updates{{\bf CT-VOS (Ours)}} & 
        \updates{ResNet-18} & 
        \updates{Kinetics (800 hrs)} & 
        \updates{{\bf 69.8}} & 
        \updates{{\bf 68.3}} & 
        \updates{{\bf 71.3}} \\
        DINO \cite{Caron_2021_ICCV} & ViT \cite{dosovitskiy2020image} & ImageNet \cite{russakovsky2015imagenet} & 71.4 & 67.9 & 74.9 \\
        {\bf CT-VOS (Ours)} & ResNet-50 & Kinetics (800 hrs) & {\bf 72.6} & {\bf 69.7} & {\bf 75.4} 
        \\
        
        \midrule
        {\bf Supervised} \\
        \midrule
        OSMN \cite{8578778} & VGG-16 & ICD (227k) & 54.8 & 52.5 & 57.1 \\
        SiamMask \cite{wang2019fast} & ResNet-50 & IVCY (2.7M) & 56.4 & 54.3 & 58.5 \\
        OSVOS\cite{Cae+17} & VGG-16 & ID (10k) & 60.3 & 56.6 & 63.9 \\
        OnAVOS \cite{DBLP:conf/bmvc/VoigtlaenderL17} & ResNet-38 & ICPD (517k) & 65.4 & 61.6 & 69.1 \\
        OSVOS-S \cite{maninis2018video} & VGG-16 & IPD (17k) & 68.0 & 64.7 & 71.3 \\
        FEELVOS \cite{voigtlaender2019feelvos} & Xception-65  & ICDY (663k) & 71.5 & 69.1 & 74.0 \\
        PReMVOS \cite{luiten2018premvos} & ResNet-101 & ICDPM (527k) & 77.8 & 73.9 & 81.8 \\
        STM \cite{oh2019video} & ResNet-50 & IDY (164k) & 81.8 & 79.2 & 84.3 \\
        SST\cite{duke2021sstvos} & ResNet-101 & IDY (164k) & 82.5 & 79.9 & 85.1 \\
        RMNet\cite{Xie_2021_CVPR} & ResNet-50 & IDY (164k) & 83.5 & 81.0 & 86.0 \\
        LCM\cite{hu2021learning} & ResNet-50 & IDY (164k) & 83.5 & 80.5 & 86.5 \\
        \bottomrule
    \end{tabularx}
    \end{center}
    \caption{State-of-the-art comparison on DAVIS-2017 validation set. Here, reference dataset notations are: I=ImageNet, V=ImageNet-VID, C=COCO, D=DAVIS, M=Mapillary, P=PASCAL-VOC, Y=YouTube-VOS.} 
    \label{table:Davis}
\end{table}

\setlength{\tabcolsep}{5.8pt}
\begin{table}[t]
    \begin{center}
        \renewcommand{\arraystretch}{1}
        \begin{tabularx}{7.9cm}{@{}l*{15}{c}c@{}}
            \toprule
            {Method} & {\it{G (\%)}} & \multicolumn{2}{c}{\it{$J$ (\%)}} & \multicolumn{2}{c}{\it{F (\%)}} \\
            \cmidrule(lr){3-4}                  
            \cmidrule(lr){5-6}
             & & {Seen} & {Unseen} & {Seen} & {Unseen} &  \\
            \midrule
            {\bf Unsupervised}\\
            \midrule
            Vid. Color \cite{vondrick2018tracking}                          & 38.9 & 43.1          & 36.6          & 38.6          & 37.4\\
            CorrFlow \cite{lai2019self}                            & 46.6 & 50.6          & 43.8          & 46.6          & 45.6\\
            MAST \cite{lai2020mast}                                & 64.2 & 63.9          & 60.3          & 64.9          &  67.7\\
            {\bf CT-VOS}                                 & {\bf 67.1}           & {\bf 67.3}         & {\bf 66.1}          & {\bf 68.4}          & {\bf 66.5}
            \\
             
            \midrule
            {\bf Supervised}\\
            \midrule
            OSMN \cite{8578778}                         & 51.2 & 60.0          & 40.6          & 60.1          & 44.0\\
            MaskTrack \cite{8099855}                    & 53.1 & 59.9          & 45.0          & 59.5          & 47.9\\
            RGMP \cite{oh2018fast}                      & 53.8 & 59.5          & 45.2          & \textendash   & \textendash\\
            OnAVOS \cite{DBLP:conf/bmvc/VoigtlaenderL17}& 55.2 & 60.1          & 46.6          & 62.7          & 51.4\\
            RVOS \cite{ventura2019rvos}                                & 56.8 & 63.6          & 45.5          & 67.2          & 51.0\\
            OSVOS \cite{Cae+17}                         & 58.8 & 59.8          & 54.2          & 60.5          & 60.7\\
            S2S \cite{5bdc315017c44a1f58a05df8}         & 64.4 & 71.0          & 55.5          & 70.0          & 61.2\\
            PReMVOS \cite{luiten2018premvos}                & 66.9 & 71.4           & 56.5          & 75.9          & 63.7\\
            AGSS-VOS \cite{Lin_2019_ICCV}               & 71.3 & 71.3          & 65.5          & 75.2          & 73.1\\
            STM \cite{oh2019video}                      & 79.4 & 79.7          & 72.8          & 84.2          & 80.9 \\
            RMNet\cite{Xie_2021_CVPR}                   & 81.5 & 82.1          & 75.7          & 85.7           & 82.4 \\
            SST\cite{duke2021sstvos}                    & 81.7 & 81.2          & 76.0          & \textendash  & \textendash \\
            LCM\cite{hu2021learning}                   & 82.0 & 82.2          & 75.7          & 86.7          & 83.4 \\
            \bottomrule
        \end{tabularx}
    \end{center}
    \caption{State-of-the-art comparison on Youtube-VOS validation set.}
    \label{table:Youtube-VOS}
\end{table}

\begin{figure*}[!t]
\centering
\includegraphics[height=8cm, width=12cm]{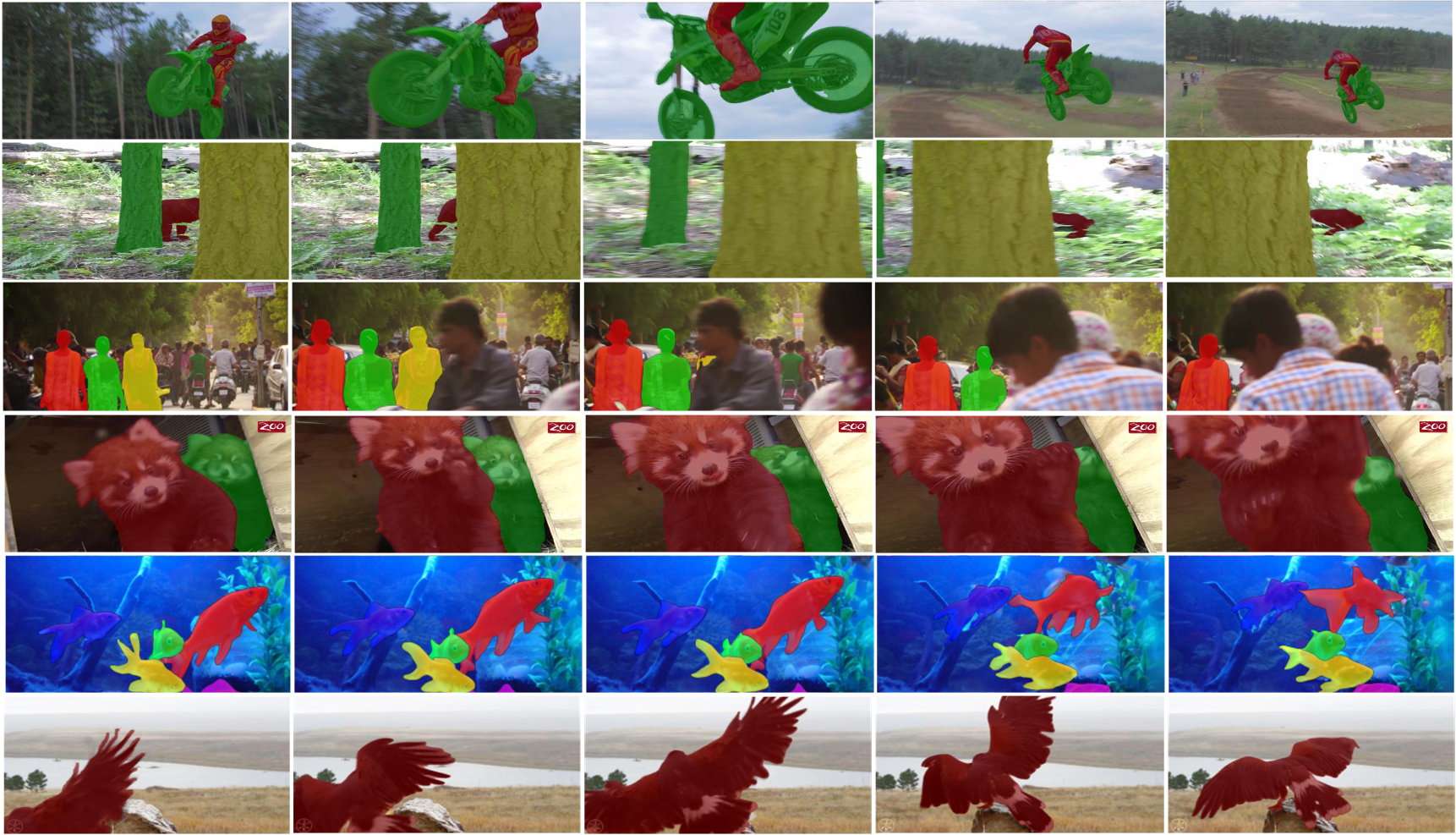}
\caption{Qualitative results of our method on DAVIS-2017 (Rows 1, 3, 5) and Youtube-VOS (Rows 2, 4, 6). The first two rows depict the ability of the model to effectively handle visual discontinuities, namely occlusion; the next two rows illustrate that our model deals well with separating objects from a background even with highly similar distractor objects; the later rows demonstrate the model's competency with generating fine-grained segmentation details for large as well as small objects.}
\label{fig:Qualitative}
\end{figure*}

\section{Experiments}

\noindent\textbf{Datasets:}
We evaluate our approach on two VOS benchmarks: DAVIS-2017 \cite{pont20172017} and Youtube-VOS \cite{xu2018youtube}. DAVIS-2017 consists of 120 videos with 60 training, 30 validation, and 30 testing sequences. Youtube-VOS comprises 4,453 annotated videos of 91 object classes.

\noindent\textbf{Training:} 
The backbone network constitutes the ResNet-18 module and additional layers that are randomly initialized. We use random crops as well as spatial and temporal flips to augment the training data. Next, the generated single clip of 8 frames is split into past frames (initial 7 frames) and the current frame ($8^{th}$ frame). In addition, we have 2 zoomed-in frames for one of the randomly chosen past frames, resulting in overall 10 frames per batch in a given epoch. The objective function used to optimize the network is discussed in detail in Section \ref{Training Losses}. For fair comparison with existing methods, we use the same training strategy as the previous self-supervised VOS works. We present the quantitative results on DAVIS-2017 validation set using backbone networks trained only on datasets mentioned under the `Dataset' column in Table \ref{table:Davis}, and the model used to report results on YouTube-VOS validation (Table \ref{table:Youtube-VOS}) is trained only using the YouTube-VOS. Our method converges on the Youtube-VOS dataset in around 400 epochs. Additionally, the model is trained using a learning rate of 0.0001 and Adam Optimizer on a Nvidia V100 GPU.

\begin{figure}
\centering
\includegraphics[height=5cm, width=8.5cm]{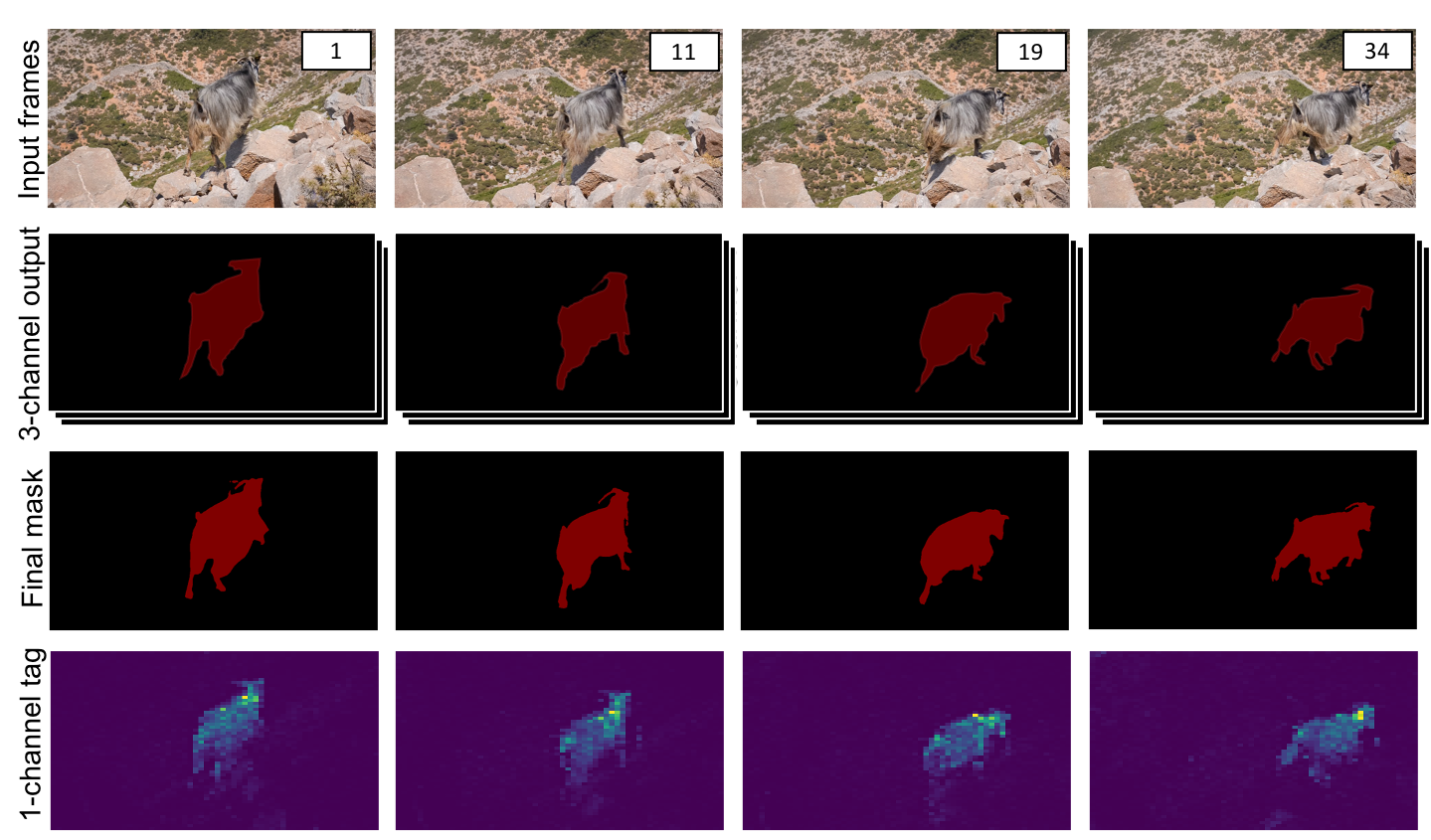}
\vspace{-3mm}
\caption{Decoder output visualizations during inference}
\vspace{-5mm}
\label{fig:Motivation}
\end{figure}

\noindent\textbf{Inference:}
Figure \ref{fig:Inference} summarizes the testing stage. Similar to the training process, we process 10 frames at-a-time during inference. Also, during inference, similar to training phase, our model generates two outputs: 3-channel output and 1-channel of tags. Tags are primarily used for enforcing object-background separation during training and not used for mask generation during inference, which is handled by a dedicated 3-channel output (2$^{nd}$-row Figure~\ref{fig:Motivation}). The predicted segmentation mask (final mask as shown in Figure~\ref{fig:Motivation}) is determined by simply averaging the generated labels from the 3-channel output. The final segmentation generated for each current frame is, then, used as input for the subsequent clip. 

\noindent\textbf{Evaluation Metrics:}
We use the following metrics to evaluate the predicted multi-object video segmentation masks in comparison to its ground truth:
\begin{itemize}
  \item[$\bullet$] Region similarity $J$: Defined by the intersection-over-union between the predicted segmentation and ground-truth mask. It provides a measure of the mislabeled pixels.
  \item[$\bullet$] Contour precision $F$: Measures the contour-based precision and recall between the contour points. It is an indicator of precision of the segmentation boundaries.
  \item[$\bullet$] Overall score $G$: Average score of $J$ and $F$ measures.
\end{itemize}

\subsection{State-of-the-art Comparison}

\begin{figure}[!t]
\centering
\includegraphics[height=10.5cm, width=8cm]{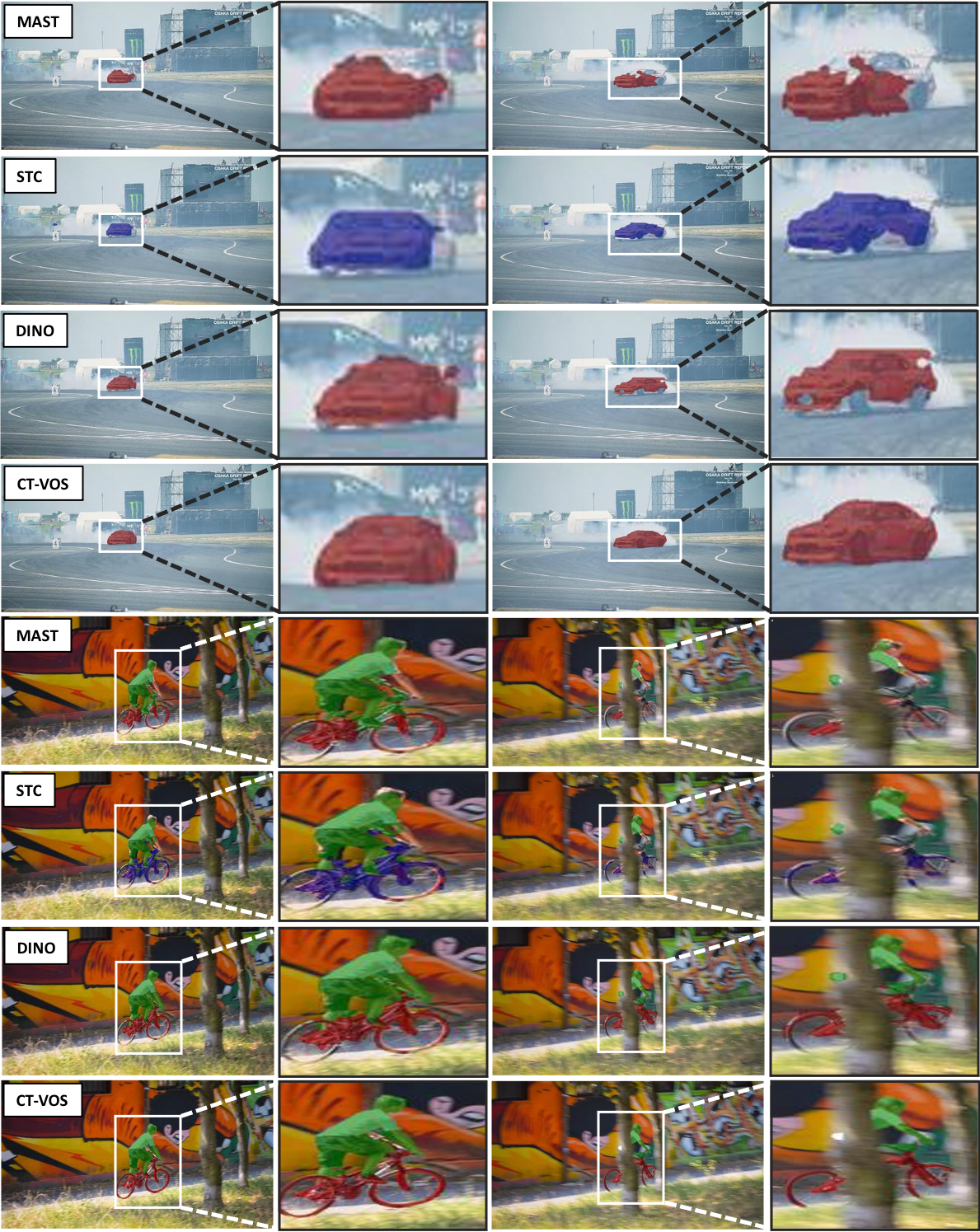}
\caption{Qualitative results on DAVIS-2017 for comparison of our method (Rows 4, 8) against existing self-supervised VOS approaches, namely MAST \cite{lai2020mast} (Rows 1, 5), STC \cite{NEURIPS2020_e2ef524f} (Rows 2, 6) and DINO \cite{Caron_2021_ICCV} (Rows 3, 7). Row 4: Ability to separate objects from highly similar surrounding distractor elements. Row 8: Generation of fine-grained segmentation details and competency to deal with occlusion.}
\label{fig:Comparison}
\end{figure}

\noindent Table \ref{table:Davis} presents our quantitative results on DAVIS-2017 validation set. Here, we observe that proposed method outperforms existing published self-supervised work on VOS. Similarly, the results on  YouTube-VOS validation set in Table \ref{table:Youtube-VOS} further affirm its ability to generate compelling segmentations. We, also, visualize the results on DAVIS-2017 and YouTube-VOS in Figure \ref{fig:Qualitative}. {\updates Furthermore, Figure \ref{fig:Comparison} highlights the improvements in segmentation that our work exhibits on DAVIS-2017 over existing self-supervised VOS methods.}

\subsection{Ablation Study}
\begin{figure}[ht]
\centering
\subfigure[Cutout-based reconstruction]{%
{\includegraphics[height=2.7cm, width=7cm]{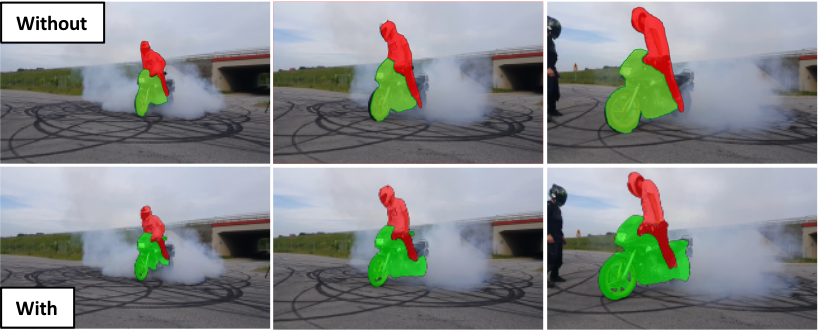}} 
\label{fig:Ablation_Cutout}}
\quad
\subfigure[Tagging loss]{%
{\includegraphics[height=2.7cm, width=7cm]{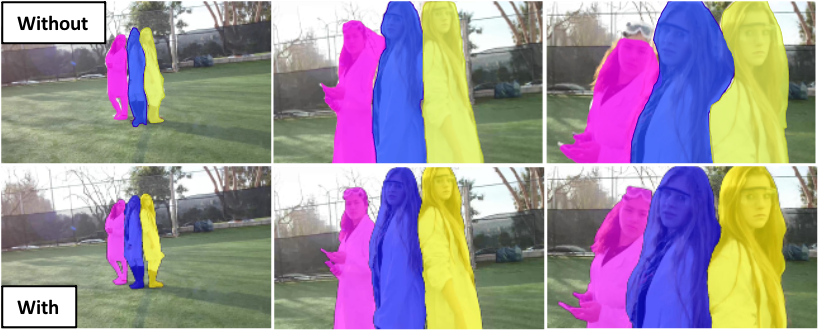}} 
\label{fig:Ablation_Tagging}}
\quad
\subfigure[Zoom-in scheme]{%
{\includegraphics[height=2.7cm, width=7cm]{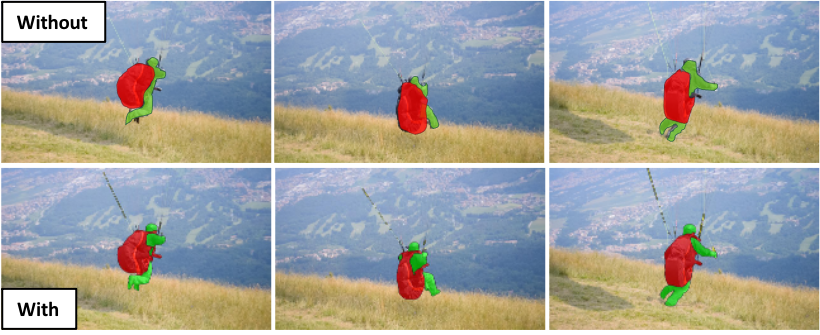}} 
\label{fig:Ablation_Zoom}}
\caption{Our qualitative results depicting the impact of each component, on DAVIS-2017, suggest that (a) Introducing cutout-based reconstruction improves model's ability to perceive spatio-temporal discontinuities (occlusion in our case). The rear-end of motorbike that is partially occluded is properly segmented by our model using  cutouts. (b) Tagging loss helps generate accurate segmentation masks for the left most person, despite the headgear blending with the surrounding pixels, thereby validating its foreground-background decoupling capabilities. (c) Fine details of paragliding backpack are captured by our zoom-in scheme.}
\label{fig:Ablation}
\end{figure}

We demonstrate the effectiveness of individual model components by adding the components sequentially. Table \ref{table:DAVIS-Ablation} summarizes the experiments performed on the DAVIS-2017 dataset with our baseline model using ResNet-18 backbone. In addition, Figure \ref{fig:Ablation} illustrates the significance of each proposed component based on the qualitative analysis performed on DAVIS-2017.

\setlength{\tabcolsep}{3.8pt}
\begin{table}
\begin{center}
    \renewcommand{\arraystretch}{1}
    \begin{tabularx}{7.2cm}{@{}l*{5}{c}c@{}}
        \toprule
        {Method} &
        {\emph{$J$\&F}} &
        {\emph{$J$\textsubscript{mean}}} & 
        {\emph{F\textsubscript{mean}}} \\
        \midrule
        Reconstruction Loss W/O Cutout& 59.7 & 58.4 & 60.9 \\
        Reconstruction Loss W Cutout  & 63.9 & 63.6 & 64.2 \\
        + Tagging Loss & 66.3 & 65.9 & 66.7 \\
        + Zoom-in Module & {\bf 68.2} & {\bf 67.1} & {\bf 69.2} \\
        \bottomrule
    \end{tabularx}
    \end{center}
    \caption{Ablation when components are added sequentially on DAVIS-2017 validation set.}
    \label{table:DAVIS-Ablation}
\end{table}

\setlength{\tabcolsep}{3.8pt}
\begin{table}
\begin{center}
    \renewcommand{\arraystretch}{1}
    \begin{tabularx}{4.9cm}{@{}l*{5}{c}c@{}}
        \toprule
        {Method} &
        {\emph{$J$\&F}} &
        {\emph{$J$\textsubscript{mean}}} & 
        {\emph{F\textsubscript{mean}}} \\
        \midrule
        Circle & 67.1 & 65.9 & 68.2 \\
        Scalene Triangle & 67.3 & 66.0 & 68.5 \\
        Rectangle & 67.9 & 66.7 & 69.1 \\
        Square & {\bf 68.2} & {\bf 67.1} & {\bf 69.2} \\
        \bottomrule
    \end{tabularx}
    \end{center}
    \caption{Impact of varying cutout shapes on DAVIS-2017 validation set.}
    \label{table:DAVIS-Ablation-Cutout-Shape}
\end{table}

\setlength{\tabcolsep}{3.8pt}
\begin{table}
    \begin{center}
    \renewcommand{\arraystretch}{1}
    \begin{tabularx}{3.9cm}{@{}l*{5}{c}c@{}}
        \toprule
        {Method} &
        {\emph{$J$\&F}} &
        {\emph{$J$\textsubscript{mean}}} & 
        {\emph{F\textsubscript{mean}}} \\
        \midrule
        {L\textsubscript{cp}} & 65.6 & 66.8 & 64.4 \\
        {L\textsubscript{rp}} & 66.7 & 65.8 & 67.5 \\
        {L\textsubscript{cp + rp}} & {\bf 68.2} & {\bf 67.1} & {\bf 69.2} \\
        \bottomrule
    \end{tabularx}
    \end{center}
    \caption{The role of pull and push components in cutout-based tagging loss.}
    \label{table:DAVIS-Ablation-Tag}
\end{table}

\setlength{\tabcolsep}{3.8pt}
\begin{table}
    \begin{center}
    \renewcommand{\arraystretch}{1}
    \begin{tabularx}{4.5cm}{@{}l*{5}{c}c@{}}
        \toprule
        {Method} &
        {\emph{$J$\&F}} &
        {\emph{$J$\textsubscript{mean}}} & 
        {\emph{F\textsubscript{mean}}} \\
        \midrule
        BCE Loss & 65.9 & 66.2 & 65.5 \\
        Tagging Loss & {\bf 68.2} & {\bf 67.1} & {\bf 69.2} \\
        \bottomrule
    \end{tabularx}
    \end{center}
    \caption{Ablation for binary cross-entropy loss v/s tagging loss.}
    \label{table:DAVIS-Ablation-Tag-Cross}
\end{table}

\setlength{\tabcolsep}{3.8pt}
\begin{table}[!t]
    \begin{center}
    \renewcommand{\arraystretch}{1}
    \begin{tabularx}{4cm}{@{}l*{5}{c}c@{}}
        \toprule
        {Method} &
        {\emph{$J$\&F}} &
        {\emph{$J$\textsubscript{mean}}} & 
        {\emph{F\textsubscript{mean}}} \\
        \midrule
        1 & {\bf 68.2} & {\bf 67.1} & {\bf 69.2} \\
        3 & 67.7 & 66.5 & 68.9 \\
        5 & 67.6 & 66.3 & 68.9 \\
       \bottomrule
    \end{tabularx}
    \end{center}
    \caption{The effect of introducing zoom-in module across multiple random past frames.}
    \label{table:DAVIS-Ablation-Zoom}
\end{table}

\noindent\textbf{Reconstruction Loss With Cutout:}
To investigate the influence of standard colorizing loss on the overall system design, we create a base model and analyze the resulting output (see Table  \ref{table:DAVIS-Ablation}). We introduce the LAB color palette to leverage the decorrelated color space, and augment the data using random crops and spatio-temporal flips in the baseline network. The qualitative results generated by this model with conventional colorizing loss exhibits significantly deteriorated object segmentation as seen in Figure \ref{fig:Ablation_Cutout}. During training, introducing cutout-based reconstruction of the images forces the model to concurrently focus on both salient as well as less salient features within the frames. Cutouts simulate occlusion-based scenarios that the model is likely to be subject to during the testing phase. It, therefore, inherently improves the generated object segmentations. Additionally, detailed analysis about shapes (refer Table \ref{table:DAVIS-Ablation-Cutout-Shape}) and sizes (refer to Table 3 from the supplementary material) of cutouts make it evident that the shape variations have inconspicuous effects on the overall performance, compared to change in the size of the cutout. The disproportionately large size of the cutout is associated with significant loss of context.

\noindent\textbf{Self-supervised Cutout-Based Tagging Loss:}
Here, we evaluate the effectiveness of the self-supervised tagging loss in our model. Adding the self-supervised tagging loss improves the overall segmentation score of the network by $\sim$2\% as depicted in Table \ref{table:DAVIS-Ablation}. In Table \ref{table:DAVIS-Ablation-Tag}, we test the model performance by introducing each component of the loss. Based on the analysis of qualitative results we notice that eliminating $L_{cp}$ causes segmentation of an individual object instance to separate into smaller disjoint clumps, while $L_{rp}$ plays a significant role in minimizing the background pixels overlapping foreground pixels of objects around edges. Here, $L_{cp}$ serves as a pulling force to bring together all the pixels associated with the background, while $L_{rp}$ pushes apart object-background pixels in close vicinity that are not related each other. {\updates Overall the tagging loss is a clustering/margin maximization-based objective. For a comprehensive ablation study, we apply binary cross-entropy loss instead of tagging loss and compare the results in Table \ref{table:DAVIS-Ablation-Tag-Cross}. We train the model using binary cross-entropy by assigning the 
cutout pixels as label 0 and the rest as 1. The quantitative results validate the ability of tagging loss to improve model performance by introducing intricate foreground-background separation. 
}

\begin{figure}[!t]
\centering
\includegraphics[height=10.5cm, width=8.5cm]{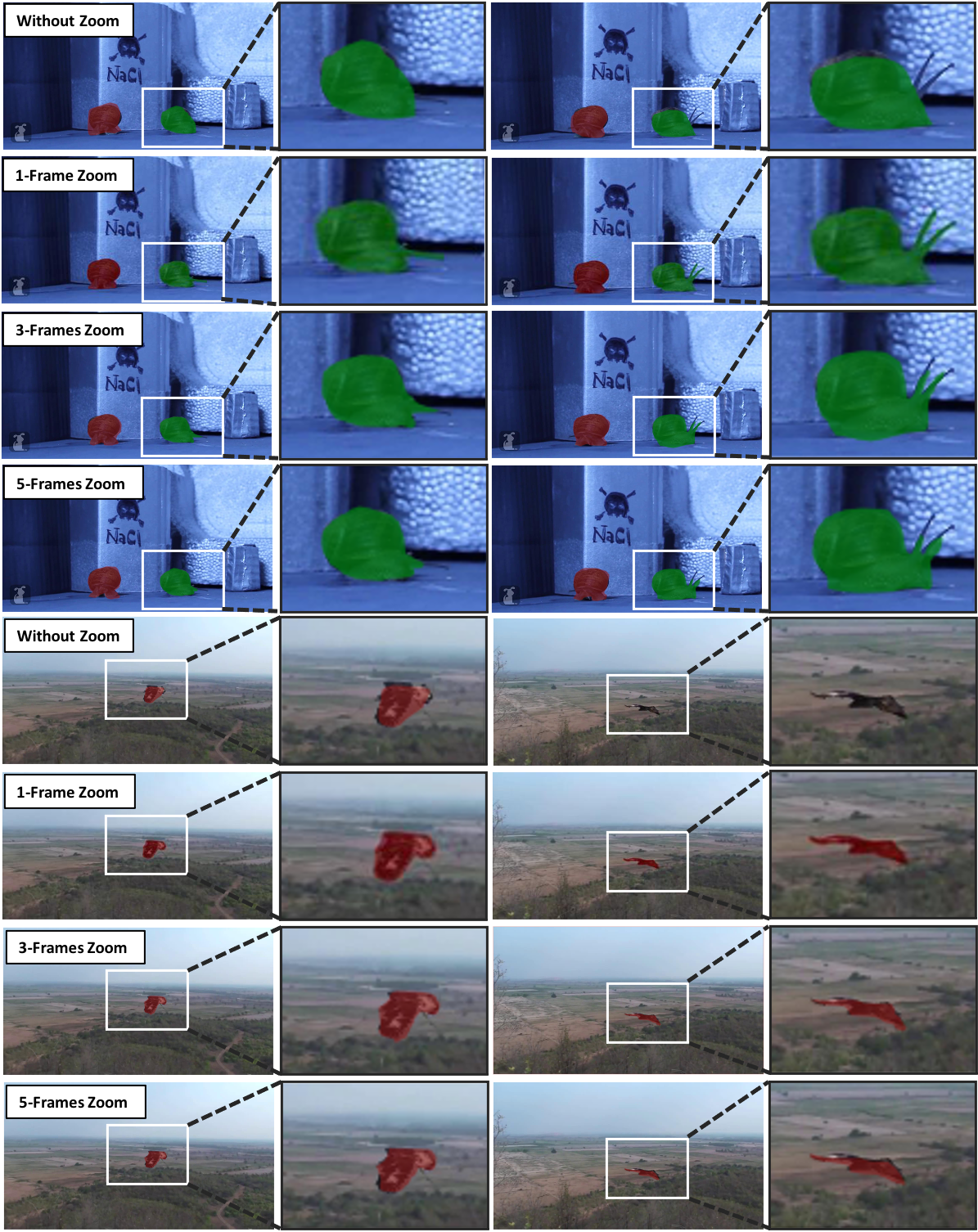}
\caption{Qualitative results displaying the ability of \textbf{zoom-in scheme} to capture fine-grained segmentation details (for quantitative results refer to Table \ref{table:DAVIS-Ablation}). In addition, we analyze the impact of introducing zoom-in module across multiple random past frames (for quantitative results refer to Table Table \ref{table:DAVIS-Ablation-Zoom}). Row 1, 2, 3, 4: Captured tentacles of the snail. Row 5, 6, 7, 8: Segmentation mask predicted for the bird's flight. Here, from the mentioned observations, it is evident that introducing the zoom-in scheme captures the finer object details, however since this component focuses on the spatial context, adding it across a single past frame or multiple past frames improves the segmentation results by a similar value.}
\label{fig:Ablation Zoom Supp}
\end{figure}

\noindent\textbf{Zoom-in Views:}
In the final ablation study, we test the importance of zoom-in views in the network. We find that the zoom-in views capture the finer structural information and thereby promotes the prediction of finer details for small objects. Table \ref{table:DAVIS-Ablation} shows the impact of the zoom-in views on the quantitative results, and we see an increase of $\sim$2\% in the overall performance. Introducing the zoom-in across multiple randomly selected past frames (see Table \ref{table:DAVIS-Ablation-Zoom}), however, fails to improve the resulting output. Since the zoom-in component focuses on exploiting the spatial information, adding it across one of the single past frame or multiple past frames improves the model outcomes by a similar value (refer to Figure \ref{fig:Ablation Zoom Supp}). The proposed zoom-in structure essentially forces the model to find clean signal from multi-scale correspondence, thereby enhancing its robustness to determining correlation. 

\section{Conclusions}

In this work, we have proposed a self-supervised video segmentation network, which exhibits improved performance. These capabilities are driven by the self-supervised cutout-based prediction and tagging loss that ensures separation of foreground-background pixels. Additionally, the zoom-in views helps by retaining the elaborate details of objects, thereby providing the context needed for segmenting smaller objects. The overall quantitative results provide the measure of the effectiveness of the network, while the ablation studies highlight the distinct contributions of each module. Our system transforms the general reconstruction-based self-supervised approach for VOS, and has the potential to be extended to more challenging segmentation and tracking tasks. 

\bibliographystyle{ACM-Reference-Format}
\bibliography{sample-base}
\end{document}